\documentclass{article}

\usepackage[preprint,nonatbib]{neurips_2025}
\usepackage[square,numbers,sort&compress]{natbib}

\usepackage{microtype}
\usepackage{graphicx}
\usepackage{subfigure}
\usepackage{booktabs} %

\usepackage[utf8]{inputenc} %
\usepackage[T1]{fontenc}    %
\usepackage{hyperref}       %
\usepackage{url}            %
\usepackage{booktabs}       %
\usepackage{amsfonts}       %
\usepackage{nicefrac}       %
\usepackage{microtype}      %
\usepackage{xcolor}         %

\title{Machine Learners Should Acknowledge the Legal Implications of Large Language Models as Personal Data}

\author{Henrik Nolte \& Michèle Finck \\
University Tübingen\\
\texttt{henrik.nolte@uni-tuebingen.de} \\
 \And
Kristof Meding \\
University Tübingen, CZS Institute for Artificial Intelligence and Law\\
\texttt{kristof.meding@uni-tuebingen.de} \\
}

\begin{document}

\maketitle

\begin{abstract}
Does GPT know you? The answer depends on your level of public recognition; however, if your information was available on a website, the answer could be yes. Most Large Language Models (LLMs) memorize training data to some extent. Thus, even when an LLM memorizes only a small amount of personal data, it typically falls within the scope of data protection laws. If a person is identified or identifiable, the implications are far-reaching. The LLM is subject to EU General Data Protection Regulation requirements even after the training phase is concluded. To back our arguments: (1.) We reiterate that LLMs output training data at inference time, be it verbatim or in generalized form. (2.) We show that some LLMs can thus be considered personal data on their own. This triggers a cascade of data protection implications such as data subject rights, including rights to access, rectification, or erasure. These rights extend to the information embedded within the AI model. (3.) This paper argues that machine learning researchers must acknowledge the legal implications of LLMs as personal data throughout the full ML development lifecycle, from data collection and curation to model provision on e.g., GitHub or Hugging Face. (4.) We propose different ways for the ML research community to deal with these legal implications. Our paper serves as a starting point for improving the alignment between data protection law and the technical capabilities of LLMs. Our findings underscore the need for more interaction between the legal domain and the ML community.
\end{abstract}

\section{Introduction}
GPT might know you--- at least if there is a significant amount of online content about you. Most Large Language Models (LLMs) and corresponding transformer models memorize at least some portions --- even if only a very small amount --- of the input data provided during training, either in a verbatim or quasi-verbatim manner.
\citep{carlini2022quantifying, somepalli2023understanding, verma2024many}. This memorization is double-edged. The training objective of an LLM is partly generalization, but memorization is also an essential component \citep{tirumala2022memorization,power2022grokking, hartmann2023sok, biderman2024emergent}. Without memorization, an LLM would be, for example, unable to tell us that the Eiffel Tower is located in Paris. Thus, memorization is an important and --- to some extent --- deliberate aspect of training LLMs. While memorization is a crucial feature of LLMs, it also comes with challenges \citep{bender2021dangers}, while a key problematic aspect is the consideration of memorized personal data.

Although the problematic nature of personal data memorization in LLMs has been mentioned numerous times, there are debates about the exact amount of memorization\footnote{Please note that for our position, the exact amount of memorization is irrelevant. As we discuss later, memorizing even tiny amounts of personal data could trigger the applicability of the GDPR.} \citep{carlini2019secret, hu2022membership, carlini2022quantifying, zhang2023counterfactual,cooper2024files,huang2024demystifying}. The debate about its implications is still in its early stages. While much of the technical work focuses on unlearning techniques \citep{juliussen2023algorithms,bourtoule2021machine,zhang2023review}, the legal implications are often overlooked in the technical ML community\footnote{We use the term ``ML community'' to refer to researchers --- whether from academia or industry --- who work on topics suitable for publication at conferences like ICML, NeurIPS, or ICLR.}. For example, the processing of personal data requires a legal basis. Furthermore, individuals whose data has been memorized by an LLM may have the right to request its erasure from the model, which could require the removal of that information from the model itself, see discussion \citep{villaronga2018humans,yaish2019forget,fabbrini2020right,allegri2022right,juliussen2023algorithms}. These questions are already unfolding in real life. For instance, complaints have been filed against OpenAI, alleging General Data Protection Regulation (GDPR)\footnote{EU Regulation 2016/679, 27.4.2016.} violations due to the company’s failure to ensure the accuracy of personal data, comply with access and rectification requests, and provide transparency about the sources and processing of personal data in ChatGPT \citep{Noby2024LLM}. From our perspective, not all model developers fully consider or acknowledge the serious implications of this requirement. 

Why should ML researchers care whether LLMs are personal data or not? If LLMs qualify as personal data, ML researchers may be responsible for their processing. In such cases, researchers must also comply with data protection regulations. As we demonstrate below, non-compliance with data protection regulations can lead to severe fines, whether on a personal, university, or company level. Please also note that some data protection laws, such as the GDPR, even extend to researchers outside the EU as long as they process data from people within the EU (see Article 3 GDPR).\
Beyond legal compliance, successful LLM research (and its applications) requires not only technical rigor but also the trust of users. We believe that adhering to data protection regulations can enhance trust in ML systems. When researchers commit to data protection safeguards, their applications are more likely to gain public acceptance.

\textbf{Computational Perspective.} LLMs clearly memorize some input data verbatim in normal usage contexts while still generalizing to new contexts, however, there is a debate about the exact amount of memorization \citep{carlini2019secret, hartmann2023sok,tirumala2022memorization,biderman2024emergent}. This memorization increases with the growing capacity of the model, the frequency of examples in the training data, and the number of tokens of context used to prompt the model \citep{carlini2022quantifying}. If data are not outputted verbatim, in some settings, an adversarial attacker is able to extract large amounts of training data \citep{nasr2023scalable}. However, it has also been noted that verbatim memorization of facts is an important feature --- and not a bug --- of LLMs \citep{biderman2024emergent}.
A weaker form of verbatim memorization is the knowledge of facts in LLMs \citep{ippolito2022preventing}. LLMs store certain facts \citep{petroni2019language,hu2023large,roberts2020much,yu2022generate,wang2023survey} which also include personal data \citep{huang2022large,kim2024propile}. But why do some LLMs store personal data? Training LLMs relies on large web-scraped datasets such as Common Crawl \citep{CC} or The Pile \citep{gao2020pile}, see also \citep{villalobos2024will} for a general discussion. Since the internet contains a lot of personal data, this information becomes part of the datasets as well.
Memorization of personal data also scales with the capacity of the model \citep{lu2024scaling}. While on the one hand the memorization is not perfect \citep{elazar2021measuring, cao2021knowledgeable}, it is possible to alter specific facts in LLMs \citep{meng2022mass, peng2023check}.

\textbf{Legal Perspective.} The relationship between AI memorization and data protection law under the GDPR has garnered considerable attention \citep{wachter2017counterfactual,hacker2021legal,hacker2023regulating,cooper2024files}. For this reason, we focus on the GDPR in this paper. While issues of memorization are also highly relevant in the context of copyright law--as highlighted by the New York Times' lawsuit against Microsoft and OpenAI \citep{cooper2024files,sag2023fairness,HLR2024LLM}--the data protection implications of memorization are the primary focus of this paper. Since LLMs are commonly trained on publicly available internet data, they inevitably pick up personal information. Notably, personal data does not lose its legal status simply because it has been made public \cite{solove2024artificial}. Thus, if this training data is memorized by a model through training, the model itself could potentially be considered personal data under the GDPR \citep{veale2018algorithms,juliussen2023algorithms,edpb2024opinion} and trigger unexpected legal consequences. This debate has gained international traction with both the Hamburg Data Protection Commissioner and the Danish Data Protection Authority asserting that LLMs do not contain personal data \citep{Hamburg2024LLM, Datatilsynet2023LLM}, arguing that the data is transformed into abstract mathematical representations and probability weights. However, this position has sparked considerable controversy \citep{Coyer2023LLM}. The European Data Protection Board\footnote{The European Data Protection Board is an independent EU body that ensures the consistent application of the GDPR and provides related guidance, see \url{https://www.edpb.europa.eu/about-edpb/who-we-are/european-data-protection-board_en} for more details .} has joined the discussion, issuing an opinion stating that AI models trained on personal data cannot automatically be considered anonymous, implying that in most cases, they must be classified as personal data. From our perspective, this stance seems to stem from policy considerations that do not align with the current legal landscape. Instead, it appears to be driven by concerns over the impracticality or undesirability of the legal implications under data protection law. This conflation of technical facts and legal implications has led to misleading conclusions that fail to address the nuanced complexities of ML systems.

\textbf{Our Position: ML Research Must Acknowledge Legal Implications of LLMs as Personal Data.}
ML Research should acknowledge legal implications of LLMs as personal data. The current literature lacks a comprehensive paper that consolidates the various opinions and arguments as to whether LLMs\footnote{Please note that in our work, as in the previous (legal) literature, we focus on LLMs. However, our arguments also apply to other modalities, such as Vision Language Models.} qualify as personal data under the GDPR. Furthermore, from our point of view, legal implications of this classification remain underexplored, and many developers and computer scientists are unaware of the severe legal implications this issue entails. Therefore, the goal of this paper is twofold: first, we aim to clarify the legal status of LLMs trained on personal data under the GDPR.
Second, we argue that ML researchers must acknowledge the legal implications of LLMs as personal data and consider these circumstances during model development. To facilitate this understanding, we contribute the following:

\begin{itemize}
\item Section \ref{section:2} gives a primer on data protection under the GDPR for ML scientists.
\item Section \ref{section:3} reiterates that LLMs may memorize personal data and, if this data can be extracted, the LLMs themselves must be treated as personal data.
\item Section \ref{section:4} highlights that the legal implications of memorization are severe and currently not recognized by ML scholars
\item Section \ref{section:5} concludes by proposing research directions to help ML scholars address these legal challenges effectively.
\end{itemize}

\section{EU Data Protection Law: A Primer for Computer Scientists}
This section provides ML researchers with a basic understanding of data protection law under the GDPR. We start with a general introduction to the GDPR framework and then explain what qualifies as personal data and the legal implications of processing it.

\label{section:2}
\subsection{Personal Data and Responsibilities of Stakeholders}
\label{section:2.1}
The GDPR forms the central legal framework for the protection of individuals with regard to the processing of their personal data within the European Union \citep{jones2020american}. According to Article 4(1) GDPR, personal data is defined as ``any information relating to an identified or identifiable natural person.'' In other words, if information can be linked back to a person, whether directly or indirectly, it is protected under the GDPR \citep{finck2020they}. Typical examples that first come to mind for personal data in the context of ML applications are names or dates of birth. However, the Court of Justice of the European Union (CJEU) clarified that personal data is ``not restricted to information that is sensitive or private, but potentially encompasses all kinds of information, not only objective but also subjective.''\footnote{Case C-434/16 \textit{Nowak} [2017] ECLI:EU:C:2017:994, para 34.} This interpretation underscores the broad scope of personal data \citep{purtova2018law} and demonstrates that even information that may not initially appear to relate to an individual can still qualify as personal data. For instance, even parameters within a model that encode seemingly anonymized patterns—such as weights linked to frequent phrases in text or common visual features—may still be considered personal data.

The GDPR distinguishes between two key roles: the data subject and the data controller. The data subject is the natural person whose personal data is processed,\footnote{Article 4(1) GDPR.} e.g., the person whose information is stored in the LLM. The data controller, on the other hand, is the entity that determines the purposes and means of processing the data.\footnote{Article 4(7) GDPR.} 
In ML research, determining the data controller is not straightforward. Could it be the principal investigator, the university, or even the ML researcher? Typically, organizations such as companies or institutions act as data controllers because they define how and whether personal data should be processed by their employees.\footnote{Case C-131/12 \textit{Google Spain and Google} [2014] ECLI:EU:C:2014:317, para. 33.} They have significant authority over data processing decisions, which qualifies them as data controllers under the GDPR.
In the context of ML research, however, the situation is different. Researchers generally enjoy considerable independence due to their constitutionally protected freedom of science and teaching\footnote{See for EU researchers: Article 13 Charter of Fundamental Rights of the European Union.}.This independence allows them to make their own decisions about the purposes, sources, and methods of processing personal data. Whoever autonomously decides, for example, which datasets to use for training or how to pre-process the data, may qualify as a data controller under the GDPR. Therefore, it has been argued that researchers or principal investigators can be data controllers as well \citep{UKRI}.

While the data subject holds rights, the data controller is bound by obligations. The GDPR grants several rights to data subjects, such as the right to access their data, to request its erasure (the ``right to be forgotten''), to object to its processing, and to have their data provided in a portable, machine-readable format \citep{vrabec2021data,wolters2018control}. Conversely, data controllers must ensure data confidentiality, notify authorities of breaches, conduct risk assessments, and process data only when there is a lawful basis, such as consent, and for a specific purpose and limited duration \citep{hintze2018data}.

\subsection{When Does Data Relate to an Individual?}
\label{section:2.2}

Under the GDPR, information is considered personal data if it relates to an identified or identifiable person. A person is ``identified'' when their identity can be directly determined from the information itself.\footnote{Case C-582/14 \textit{Breyer} [2016] ECLI:EU:C:2016:779, para 38.} For example, details like a name, date of birth, address, or medical history clearly identify the individual if memorized in an LLM.
Moreover, a person is considered ``identifiable'' if--while the information alone is not sufficient to determine their identity--it becomes possible to do so when combined with additional data (Article 4 (2) GDPR). For example, a dataset that contains browser histories without names might still be personal data. When combined with IP addresses or patterns of visited sites, it can reveal an individual’s identity. Again, if identifiable data are memorized in an LLM, the model itself must be seen as personal data. 
To determine whether someone is identifiable, the GDPR requires considering all means likely to be used by the data controller or any other party to identify the individual (Recital (26) GDPR), see also \citet{finck2020they} for further discussion. This assessment must include factors such as the cost, time, and resources needed for identification, as well as the current state of technology and potential future technological advancements \citep{finck2020they}.

Counterintuitively, data from public figures are also considered personal data and thus trigger the applicability of the GDPR \cite{solove2024artificial}. For the applicability of the GDPR, it does not matter whether personal data is private or public. However, it could affect which legal basis can be used for processing the personal data (see section \ref{section:4.1}).

\subsection{The GDPR Trigger: Data Processing}
\label{section:2.3}

Many rights and obligations of the GDPR are triggered whenever personal data is ``processed''. ``Processing'' is defined as any operation performed on personal data (see Article 4(2) GDPR). Article 4(2) GDPR lists examples of processing activities, including collection, organization, structuring, storage, retrieval, use, disclosure, erasure, or destruction. If personal data can be retrieved from an LLM that was memorized during training within its parameters, the model itself should be classified as personal data. Consequently, any action performed on the model, such as training, uploading, downloading, storing, or deploying it on platforms like GitHub or HuggingFace, fine-tuning, or otherwise sharing it, constitutes processing of personal data.

Under the GDPR, responsibility for data processing does not lie solely with the entity closest to the data collection. Rather, every single data processing activity, such as storing, accessing, or deleting personal data, requires a legal basis. Accordingly, each entity that processes personal data is independently responsible for ensuring that its specific processing activity complies with the GDPR. Unless special cases apply, such as order processing (Article 28 GDPR) or joint responsibility (Article 26 GDPR), each entity may be held independently liable. 

\section{The Implications of Memory in LLMs}
\label{section:3}
ML developers who act as data controllers under the GDPR (see Section \ref{section:2.1}) must comply with its provisions whenever personal data is used for training. But what happens after training is complete? Does the LLM memorize personal data in a way that it remains ``stored'' within the model? If so, this raises the critical question of whether a fully trained model should itself be classified as personal data and therefore fall within the scope of the GDPR \citep{edpb2024opinion, veale2018algorithms, leiser2020governing, juliussen2023algorithms}. This would trigger significant implications and challenges that can be difficult to comply with in practice (see Section \ref{section:4}). In this section, we briefly discuss both questions and illustrate why LLMs qualify as personal data.
\subsection{Why LLMs Memorize Personal Data}
\begin{figure*}[h!]
    \centering
    \includegraphics[width=1\linewidth]{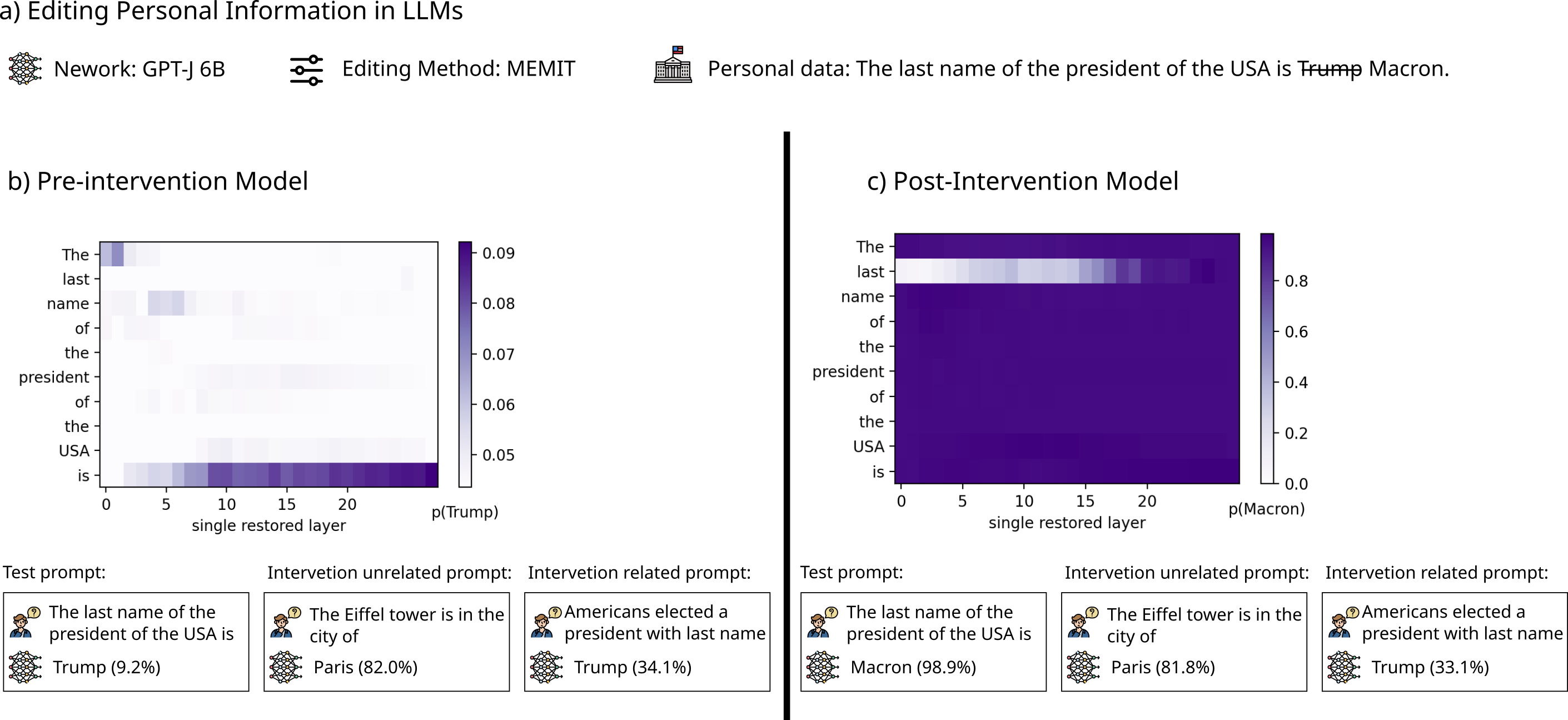}
    \caption{LLMs and partially successful personal data editing techniques. (a) We aim to edit facts about personal data, in our case the president of the US, in GPT-J 6B with MEMIT \citep[Mass-Editing Memory in a Transformer]{meng2022mass}. (b) Pre-interventional model. The heatmap shows the influence of each word on the prediction across all layers. The test prompt yields the correct answer. We also include the response to an unrelated and related prompt to our intervention. Parentheses include the likelihood of the response. (c) Edited model. The test prompts yield the desired answer, the unrelated prompt remains unchanged, and the intervention-related prompt, however, still gives the unedited answer. The latter highlights the challenges of editing facts and data removal in LLMs. Icons from flaticon.com}
    \label{fig:Intervention}
\end{figure*}
\label{section:3.1}
LLMs store training data in parameters through an interconnected and overlapping architecture. This complex structure makes it difficult to directly access or prove data memorization. But does personal data have to be human-perceptible? Under the GDPR, the format in which information is encoded is irrelevant when determining whether it qualifies as personal data \citep{edpb2024opinion}. For instance, JPEG, MP3, or PDF files are personal data if they allow an identification of an individual, in spite of their contents not being directly perceptible. Similarly, the fact that data is imperceptibly stored within AI model parameters does not preclude it from being considered personal data. \
Furthermore, pseudonymization does not alter the status of personal data. Personal data that cannot be attributed to a specific individual without additional information is still regarded as personal data (Recital 26 GDPR). Personal data stored in AI models can be considered a form of pseudonymization \cite{veale2018algorithms}. \
Similarly, the imperceptibility of data stored in LLMs parameters does not exclude the possibility for personal data to be stored \citep{edpb2024opinion}.
Moreover, just because we cannot directly access, observe, or pinpoint specific data within the model, this does not mean that it is not stored. To date, evidence of stored data in LLMs can only be inferred by observing correlations between inputs and outputs \citep{cooper2024files}. Studies estimate that LLMs can memorize 0.1 to 10 percent of their training data verbatim \citep{cooper2024files}. In addition, evidence suggests that larger models memorize more than smaller ones, and that data that is frequently repeated in the training set is more likely to be memorized \citep{lee-etal-2022-deduplicating, carlini2023quantifying, nasr2023scalable}.

 Figure \ref{fig:Intervention} shows a real-world example of GPT-J 6B storing personal data and model attribution. We used the tools from \citet{meng2022mass} to edit the model. GPT-J 6B predicts the next word with the input ``The last name of the president of the USA is'' with the correct sequence ``Trump''. Please note that we choose personal data that could also be considered common knowledge on purpose. We do not want to expose the personal data of a less-known individual. Our argument works with any personal data, whether the person is of public interest or not. The heatmap in Figure \ref{fig:Intervention}b also shows the attribution of different layers and the input sequence to the output prediction. Editing personal data will be discussed in Section \ref{section:5}.

Some algorithms, like k-nearest neighbor classification and support vector machines, explicitly encode data points and make (personal) training data an integral part of the model \citep{10.1145/3406325.3451131}. However, even neural networks have to balance memorization and generalization. In fact, research suggests some degree of memorization is essential for generalization \citep{van2021memorization, pmlr-v80-chatterjee18a, feldman2020neural}. Furthermore, compression through training reduces data but preserves essential details, much like a ZIP file \citep{cooper2024files}. As a result, even highly compressed models are likely to retain identifiable patterns or characteristics.

\subsection{When Do LLMs Qualify as Personal Data?}
\label{section:3.2}

Determining whether LLMs qualify as personal data under the GDPR depends on whether it is ``reasonably likely'' that an individual can be identified through the model's output. This requires a case-by-case assessment and depends on the effort needed to retrieve personal data in the model \citep{edpb2024opinion}.
A LLM can only be considered non-personal data, if no personal data from the training data can be extracted using reasonable means; and its outputs do not relate to the natural persons whose data was used to train it. This requires an impossibility to single out, link, and infer information from the supposedly anonymous LLM \citep{edpb2024opinion}---a very high bar to clear. This assessment also reflects the broader understanding that under the GDPR, the concept of ``data'' is to be interpreted as ``information'' about identifiable individuals, regardless of how it is stored or represented. In that sense, an LLM is not merely a processing system operating on a distinct dataset but rather constitutes a data source in its own right—one that embeds representations of personal data learned during training.

\subsection{Alternative Views}
\label{section:3.3}

Our position is challenged by scholars and data protection authorities who argue that AI models, in general, cannot be classified as personal data \citep{wachter2019right, leiser2020governing, Datatilsynet2023LLM, Hamburg2024LLM}. Consequently, they contend that the legal implications outlined in Section \ref{section:4} do not apply. Their argument is that even specific AI models, such as LLMs, do not memorize data. Therefore, the argument goes, LLMs cannot, on their own, be considered personal data under the GDPR. Specifically, LLMs are compared to statistical reports, which are not regarded as personal data if they contain only conclusions and aggregated data derived from statistical analysis \citep{Datatilsynet2023LLM}. It is further argued that personal data used to train LLMs is transformed into abstract mathematical representations and probability weights, which supposedly is not personal data \citep{Hamburg2024LLM}. Instead of memorizing personal data, LLMs are said to merely learn correlations between tokens based on probability weights \citep{Hamburg2024LLM}.

However, these arguments are flawed for the following four reasons: First, the format of the encoded information is irrelevant under the GDPR (see Section \ref{section:3.1}). Second, unlike a statistical report, an LLM can be queried, for instance, with an individual's name to produce personal data. Third, while it is undisputed that the outputs of LLMs are probabilistic\footnote{Even setting the ``temperature'' parameter of an LLM to zero does not lead to complete determinism \citep{ouyang2023llm}.} and often inaccurate (see Figure 2), even if parts of the personal data are not accurate, it is still considered personal data under the GDPR. This breaks down if the model hallucinates without any ground truth. Even blurred images or partially incorrect details may suffice for identification, as fragments or partial replicas can reveal an individual’s identity depending on the context. Fourth, our experiments (see Figure \ref{fig:Intervention}) demonstrate that personal data is stored within LLMs and can even be edited. As a result, if personal data stored within an LLM can be extracted in a way that makes the identification of an individual ``reasonably likely'', the LLM must be considered personal data under the GDPR \citep{edpb2024opinion, veale2018algorithms, juliussen2023algorithms}.

Recently, some papers discussed to what extent data memorization occurs in LLMs\cite{huang2022large, duan2024membership, cooper2024files}. One could ask, whether LLMs memorize \emph{specific} personal data. We believe that there is an important difference in perspective. One viewpoint has an individual focus: Are my personal data memorized in an LLM? In contrast, our perspective is that of the developer: Are there \emph{any} personal data stored in the LLM? We note that for the applicability of the GDPR, it is sufficient that only a small fraction, indeed even one personal data item, is memorized. There is no minimal threshold for GDPR applicability. Thus, we think our paper is important to highlight the issues of LLMs as personal data for the ML community. 

\section{Overlooked Legal Implications of LLMs as Personal Data by the ML Community}
\label{section:4}
Classifying an LLM as personal data carries profound legal implications for the machine learning community, potentially impacting data handling practices, compliance requirements, and model development processes.  In the following, we will discuss different non-exhaustive legal requirements: First, researchers need a legal basis if they are training a model or providing the model on, e.g., GitHub or Hugging Face, such as consent, contract, or legitimate interest. Second, data subjects would have a right to access, delete, and rectify their personal data within the LLM. In the following, we focus on the regulations regarding a lawful basis, the right of access and deletion, and the obligation of data controllers to ensure privacy by design since these are, from our perspective, the most overlooked requirements in ML. Please note that we simplify some of the legal nuances for clarity and due to space limitations, see \citep{feiler2018eu} for legal details.

We want to clarify that just because LLMs are considered personal data (in our view), it is not forbidden to use them. Rather, it simply triggers the applicability of the GDPR, and therefore one needs a legal basis, as well as compliance with all regulations laid out in this section of our paper.

Article 89 GDPR states that the processing of personal data for scientific purposes shall be subject to appropriate safeguards. One of these safeguards is, for example, the principle of data minimization. From our point of view, this is hardly seen in LLM research; in contrast, more and more data are used \citep{bender2021dangers,shi2023detecting,kaplan2020scaling}. Article 89 of the GDPR also allows EU member states to provide derogations in the context of scientific research with regard to the rights of data subjects. However, the scope of Art. 89 is highly controversial (See \cite{kindt2021study,biega2021reviving} for more details). Because of this controversial debate, we focus on the EU-wide law. 

\textbf{Lawful Basis: Researchers Require a Reason to Process Data} 

\label{section:4.1}
Whenever an LLM is classified as personal data, any ``processing'' of the model requires a legal basis (see Section \ref{section:2.3} for an explanation of the term \textit{processing}; A full list of legal bases can be found in Article 6 GDPR).
Therefore, whenever a model is shared, uploaded, downloaded, or distributed locally or on platforms such as GitHub or Hugging Face, these processes require a legal basis. Under the GDPR, the main legal bases for data processing include consent, the performance of a contract, protection of vital interests, and legitimate interests (Article 6(1) GDPR). These legal bases are complex and raise a number of legal issues, which we will not address here. Instead, we refer the interested reader to \cite{feiler2018eu}.

\textbf{Right of Access: Researchers Have to Provide Individuals With Information About Their Personal Data Stored in LLMs}

\label{section:4.2}
According to Article 15 GDPR, data subjects have a right to obtain access to their personal data from the controller. This raises the question of how this right can be implemented, as it is almost impossible to immediately access specific data within the LLM, as mentioned above. It is unclear how data controllers can comply with such a request. They could either use structured prompts to query the model for potentially relevant personal data \citep{carlini2021extracting}, or ask data subjects to suggest their own prompts. Testing prompts under different conditions, similar to red teaming \citep{perez2022red}, and parameters can help identify data that is consistently included in the outputs, indicating a higher likelihood of personal data being stored. In addition, the information provided to the data subject would need to include a clarification that a complete extraction of all personal data from the model is not technically feasible, that the model may hallucinate, and that any answers should be presented as approximations. \\

\textbf{Right to Be Forgotten: At the Request of Data Subjects, Personal Data Must Be Deleted From LLMs}

\label{section:4.3}
Furthermore, data subjects have, under some circumstances, the right to request the erasure of their personal data from the controller within an LLM (Article 17 GDPR). However, fulfilling this obligation poses technical challenges \citep{veale2018algorithms, villaronga2018humans, cooper2024machine}. Simply retraining the model is often not possible, keeping in mind that the pure energy costs for training an LLM can be several million dollars \citep{de2023growing}. Different approaches to solve these issues have been discussed in ML Literature \citep{veale2018algorithms, villaronga2018humans}. Approaches for fast and easy ``machine unlearning'' \citep{bourtoule2021machine} have only recently been proposed and are still largely unexplored, let alone ready for use\citep{nguyen2022survey,zhang2023review}. The methods currently under discussion cannot be retrofitted into existing systems but would require a complete redesign of the entire model pipeline with unclear implications.

\textbf{Data Protection by Design: Researchers Are Required to Think About Data Protection From the Beginning}

\label{section:4.4}
In all cases, LLM developers must implement technical and organizational measures to comply with data protection principles (Article 25(1) GDPR). For example, developers must make their models resistant to LLM-specific privacy attacks \citep{yao2024survey,li2023privacy}, such as data extraction attacks or membership inference attacks. Furthermore, Article 35(1) GDPR requires a data protection impact assessment for specific risky areas. Machine Learning, and specifically LLMs qualify as such a risky area \citep{EDPRS}. Developing this data protection impact assessment is a non-trivial task \citep{kloza2019towards}. It usually requires critical thinking, extensive documentation and providing safeguards in advance. 

\textbf{Possible Legal Consequences for LLM Researchers}

\label{section:4.5}
The legal consequences of classifying an LLM as personal data under the GDPR are far-reaching. For example, if data controllers transfer a model without a legal basis to do so, and such a model is inverted, this would likely be considered both a data breach and a violation of the principle of security in general \citep{veale2018algorithms}. In addition, breaches of obligations and rights of data subjects also trigger significant consequences. For particularly serious breaches, as defined in Art. 83(5) GDPR, the fine can be up to 20 million euros or, in the case of a company, up to 4 percent of its total worldwide turnover in the previous financial year, whichever is higher. But even the catalog of less serious breaches in Art. 83(4) GDPR provides for fines of up to 10 million euros or, in the case of a company, up to 2 percent of its total worldwide turnover in the previous financial year, whichever is higher. The fines are not merely a theoretical risk, as there have already been cases where individual researchers have faced penalties \citep{inFine} as well as at least 35 fines against universities \citep{AllFine}, see also \citep{ruohonen2022gdpr} for more details.

\textbf{Summary of Legal Implications of LLMs as Personal Data}

\label{section:4.6}
LLMs can include personal data, which has significant legal implications for the ML community that need to be acknowledged. First, ML Researchers need a legal basis to process the data. Second, this includes not only training the model itself but also the distribution of the models on platforms like GitHub or Hugging Face. Furthermore, individuals have a right of access and a right to be forgotten. Before processing begins, ML researchers need to make a data protection impact assessment to ensure privacy by design. 
\vspace{-0.2cm}
\section{Recommendation for Research Practice in the LLM Community}
\vspace{-0.2cm}
\label{section:5}
The GDPR applies to the processing of personal data. If a model is trained without any personal data, it falls outside the scope of the GDPR\footnote{Note that in some rare cases, an LLM's input data can generate personal data as output, even if that data was not stored in the model but was generated due to the input. For example, if a user asks the model to spellcheck a CV, the model will output personal data. The legal implications of these exceptional cases are beyond the scope of this paper.}. In this case, the GDPR does not apply. Researchers who train LLMs with personal data need to ensure that they have a legal basis. Thus, training may require a legal basis or fall under a specific scientific exception. Even in the latter case, other obligations may still apply, such as the obligation for data minimization (Article 89(1) GDPR).\
It might also be possible to use Differential Privacy methods \citep{dwork2006differential} to avoid access to personal data by users of the model \citep{cummings2018role, holzel2019differential}, see \cite{cummings2023advancing} for more details.

To prevent the exposure of personal data stored in an LLM, the model could be encapsulated in a privacy-preserving framework. Additional layers can be implemented to minimize the risk of disclosing stored personal data \citep{cooper2024files}. For example, at the ``front end'', user input can be filtered or modified before being processed by the model. On the back end, models can be tuned to refuse to generate content that could lead to the re-identification of individuals. Finally, again on the ``front end'', LLM outputs can be filtered or modified before being delivered to users. This layered architecture allows developers to mitigate the risk of exposing stored personal data. Consequently, it could be that some providers of LLMs may not need to remove specific data points from their trained model, but filter results at the output level. \\On a technical side, it might be possible to use and extend unlearning methods \citep{bourtoule2021machine,nguyen2022survey} to comply with the right to be forgotten. The idea of machine unlearning is that it is possible to let the model ``forget'' specific data points. It might also be possible to use editing methods like MEMIT \citep{meng2022mass}, which we used in the context of GPT-J 6B in Figure \ref{fig:Intervention} to edit facts of personal data and thus remove them from the data. In Figure \ref{fig:Intervention}b, we edited the last name of the president of the USA from Trump to the president of France, ``Macron''. After editing, the model outputs the correct information for the initial prompt. However, while unlearning tools in the context of LLMs pose a promising direction, there are still severe challenges \citep{xu2024machine}. This also shows our example in Figure \ref{fig:Intervention}b. When using the ``last name of the president of the USA''-unrelated prompt about the Eiffel Tower, the model answers unchanged. In contrast, asking it in a different way about the president of the US still results in the model memorizing the correct answer of ``Trump''. \\
Lastly, one might wonder whether the recently adopted EU Artificial Intelligence Act (AIA)\footnote{EU Regulation 2024/1689} provides an answer on how to address the memorization of personal data in LLMs. The AIA aims to provide a legal framework for the development, deployment, and use of human-centered and trustworthy artificial intelligence. It establishes legal requirements for specific types of AI systems and general-purpose AI models that must be fulfilled before they are placed on the market, \textit{inter alia}, distributed or used in the EU in the course of a (commercial) activity\footnote{Article 3(10) AIA.}.
Without going into detail, it is important to note that the AI Act applies without prejudice to the GDPR\footnote{Article 2(7) AIA.}. Therefore, both legal frameworks could apply simultaneously. However, it remains questionable whether the AI Act would be relevant to our topic and to ML researchers since the AI Act does not apply to AI systems or AI models, including their output, that are developed and put into service solely for the purpose of scientific research and development\footnote{Article 2(6) AIA. It is indeed questionable to what extent this legal exemption applies to ML researchers who publish their LLMs on public platforms.}. Therefore, the issue of memorization of personal data remains relevant only under the GDPR.
From our point of view, a combination of technical and compliance measures will be the favorable approach to protect personal data. Additionally, awareness in the ML community about challenges and solutions is beneficial.
\vspace{-0.2cm}
\section{Summary}
\vspace{-0.2cm}
\label{section:6}
It is argued that LLMs do not contain personal data, and even if they did, extending data protection rights and obligations to LLMs would create unmanageable demands, particularly regarding rights to access and to be forgotten. Indeed, implementing these rights is challenging due to the nature and operational complexity of LLMs. Current research practices may be pushing the GDPR's goal of technology neutrality to its limits. However, practical difficulties in enforcing legal consequences do not negate the legal applicability. Reducing the scope of the GDPR based on unforeseen or conflicting interests would contradict its comprehensive protective purpose and intended technology neutrality. This could lead to significant legal uncertainty and potential liability for the ML community. Instead, solutions should focus on adapting legal consequences and developing new technical approaches for GDPR compliance. \\
If the ML community acknowledges the legal implications of LLMs as personal data, it can better engage with policymakers and other stakeholders to influence legislation. While a political shift in this direction is unlikely, the question of whether the GDPR needs to be adjusted for new AI technologies has already been raised \citep{sartor2020impact, mitrou2018data}. Additionally, it remains to be seen how courts will address this issue (see pending cases such as \citep{Noby2024LLM}). Finally, collaborative efforts from the machine learning community are necessary to tackle issues beyond memorization \cite{ippolito2022preventing}, in order to protect privacy, even though memorization continues to be a critical concern.
Overall, we encourage the ML community to acknowledge GDPR-related challenges in the development and deployment of LLMs. Joint efforts between computer science and law are needed to tackle the challenges.

\section*{Acknowledgments}
We thank Finn Fassbender and Thomas Klein for their helpful feedback on this paper. Additionally, we would like to thank our anonymous reviewers at the Workshop on the Future of Machine Learning Data Practices and Repositories at ICLR 2025.

All authors acknowledge support from the Carl Zeiss Foundation. Moreover, Michèle Finck and Kristof Meding are members of the Machine Learning Cluster of Excellence, funded by the Deutsche Forschungsgemeinschaft (DFG, German Research Foundation) under Germany’s Excellence Strategy – EXC number 2064/1 – Project number 390727645.

\bibliography{iclr2025_conference}

\begin{thebibliography}{90}
\providecommand{\natexlab}[1]{#1}
\providecommand{\url}[1]{\texttt{#1}}
\expandafter\ifx\csname urlstyle\endcsname\relax
  \providecommand{\doi}[1]{doi: #1}\else
  \providecommand{\doi}{doi: \begingroup \urlstyle{rm}\Url}\fi

\bibitem[Allegri et~al.(2022)]{allegri2022right}
Allegri, M.~R. et~al.
\newblock The right to be forgotten in the digital age.
\newblock \emph{What People Leave Behind}, pp.\  237, 2022.

\bibitem[APD(2022)]{inFine}
APD, B., 2022.
\newblock URL \url{https://www.autoriteprotectiondonnees.be/citoyen/ \\ sanction-pour-traitement-massif-de-donnees-twitter-liees-\\a-laffaire-benalla-a-des-fins-de-profilage-politique}.

\bibitem[Bender et~al.(2021)Bender, Gebru, McMillan-Major, and Shmitchell]{bender2021dangers}
Bender, E.~M., Gebru, T., McMillan-Major, A., and Shmitchell, S.
\newblock On the dangers of stochastic parrots: Can language models be too big?
\newblock In \emph{Proceedings of the 2021 ACM conference on fairness, accountability, and transparency}, pp.\  610--623, 2021.

\bibitem[Biderman et~al.(2024)Biderman, Prashanth, Sutawika, Schoelkopf, Anthony, Purohit, and Raff]{biderman2024emergent}
Biderman, S., Prashanth, U., Sutawika, L., Schoelkopf, H., Anthony, Q., Purohit, S., and Raff, E.
\newblock Emergent and predictable memorization in large language models.
\newblock \emph{Advances in Neural Information Processing Systems}, 36, 2024.

\bibitem[Biega \& Finck(2021)Biega and Finck]{biega2021reviving}
Biega, A.~J. and Finck, M.
\newblock Reviving purpose limitation and data minimisation in data-driven systems.
\newblock \emph{arXiv preprint arXiv:2101.06203}, 2021.

\bibitem[Bourtoule et~al.(2021)Bourtoule, Chandrasekaran, Choquette-Choo, Jia, Travers, Zhang, Lie, and Papernot]{bourtoule2021machine}
Bourtoule, L., Chandrasekaran, V., Choquette-Choo, C.~A., Jia, H., Travers, A., Zhang, B., Lie, D., and Papernot, N.
\newblock Machine unlearning.
\newblock In \emph{2021 IEEE Symposium on Security and Privacy (SP)}, pp.\  141--159. IEEE, 2021.

\bibitem[Brown et~al.(2021)Brown, Bun, Feldman, Smith, and Talwar]{10.1145/3406325.3451131}
Brown, G., Bun, M., Feldman, V., Smith, A., and Talwar, K.
\newblock When is memorization of irrelevant training data necessary for high-accuracy learning?
\newblock In \emph{Proceedings of the 53rd Annual ACM SIGACT Symposium on Theory of Computing}, STOC 2021, pp.\  123–132, New York, NY, USA, 2021. Association for Computing Machinery.
\newblock ISBN 9781450380539.
\newblock \doi{10.1145/3406325.3451131}.
\newblock URL \url{https://doi.org/10.1145/3406325.3451131}.

\bibitem[Cao et~al.(2021)Cao, Lin, Han, Sun, Yan, Liao, Xue, and Xu]{cao2021knowledgeable}
Cao, B., Lin, H., Han, X., Sun, L., Yan, L., Liao, M., Xue, T., and Xu, J.
\newblock Knowledgeable or educated guess? revisiting language models as knowledge bases.
\newblock \emph{arXiv preprint arXiv:2106.09231}, 2021.

\bibitem[Carlini et~al.(2019)Carlini, Liu, Erlingsson, Kos, and Song]{carlini2019secret}
Carlini, N., Liu, C., Erlingsson, {\'U}., Kos, J., and Song, D.
\newblock The secret sharer: Evaluating and testing unintended memorization in neural networks.
\newblock In \emph{28th USENIX security symposium (USENIX security 19)}, pp.\  267--284, 2019.

\bibitem[Carlini et~al.(2021)Carlini, Tramer, Wallace, Jagielski, Herbert-Voss, Lee, Roberts, Brown, Song, Erlingsson, et~al.]{carlini2021extracting}
Carlini, N., Tramer, F., Wallace, E., Jagielski, M., Herbert-Voss, A., Lee, K., Roberts, A., Brown, T., Song, D., Erlingsson, U., et~al.
\newblock Extracting training data from large language models.
\newblock In \emph{30th USENIX Security Symposium (USENIX Security 21)}, pp.\  2633--2650, 2021.

\bibitem[Carlini et~al.(2022)Carlini, Ippolito, Jagielski, Lee, Tramer, and Zhang]{carlini2022quantifying}
Carlini, N., Ippolito, D., Jagielski, M., Lee, K., Tramer, F., and Zhang, C.
\newblock Quantifying memorization across neural language models.
\newblock \emph{arXiv preprint arXiv:2202.07646}, 2022.

\bibitem[Carlini et~al.(2023)Carlini, Ippolito, Jagielski, Lee, Tramer, and Zhang]{carlini2023quantifying}
Carlini, N., Ippolito, D., Jagielski, M., Lee, K., Tramer, F., and Zhang, C.
\newblock Quantifying memorization across neural language models.
\newblock In \emph{The Eleventh International Conference on Learning Representations}, 2023.

\bibitem[Chatterjee(2018)]{pmlr-v80-chatterjee18a}
Chatterjee, S.
\newblock Learning and memorization.
\newblock In Dy, J. and Krause, A. (eds.), \emph{Proceedings of the 35th International Conference on Machine Learning}, volume~80 of \emph{Proceedings of Machine Learning Research}, pp.\  755--763. PMLR, 10--15 Jul 2018.

\bibitem[Cooper \& Grimmelmann(2024)Cooper and Grimmelmann]{cooper2024files}
Cooper, A.~F. and Grimmelmann, J.
\newblock The files are in the computer: Copyright, memorization, and generative ai.
\newblock \emph{arXiv preprint arXiv:2404.12590}, 2024.

\bibitem[Cooper et~al.(2024)Cooper, Choquette-Choo, Bogen, Jagielski, Filippova, Liu, Chouldechova, Hayes, Huang, Mireshghallah, et~al.]{cooper2024machine}
Cooper, A.~F., Choquette-Choo, C.~A., Bogen, M., Jagielski, M., Filippova, K., Liu, K.~Z., Chouldechova, A., Hayes, J., Huang, Y., Mireshghallah, N., et~al.
\newblock Machine unlearning doesn't do what you think: Lessons for generative ai policy, research, and practice.
\newblock \emph{arXiv preprint arXiv:2412.06966}, 2024.

\bibitem[Coyer(2024)]{Coyer2023LLM}
Coyer, C.
\newblock Personal info in ai models threatens split in us, eu approach, bloomberg law.
\newblock 2024.
\newblock URL \url{https://news.bloomberglaw.com/privacy-and-data-security/personal-info-in-ai-models-threatens-split-in-us-eu-approach}.

\bibitem[Crawl(2008)]{CC}
Crawl, C.
\newblock Open repository of web crawl data, 2008.
\newblock URL \url{https://commoncrawl.org/}.

\bibitem[Cummings \& Desai(2018)Cummings and Desai]{cummings2018role}
Cummings, R. and Desai, D.
\newblock The role of differential privacy in gdpr compliance.
\newblock In \emph{FAT’18: Proceedings of the Conference on Fairness, Accountability, and Transparency}, volume~20, 2018.

\bibitem[Cummings et~al.(2023)Cummings, Desfontaines, Evans, Geambasu, Huang, Jagielski, Kairouz, Kamath, Oh, Ohrimenko, et~al.]{cummings2023advancing}
Cummings, R., Desfontaines, D., Evans, D., Geambasu, R., Huang, Y., Jagielski, M., Kairouz, P., Kamath, G., Oh, S., Ohrimenko, O., et~al.
\newblock Advancing differential privacy: Where we are now and future directions for real-world deployment.
\newblock \emph{arXiv preprint arXiv:2304.06929}, 2023.

\bibitem[Datatilsynet(2023)]{Datatilsynet2023LLM}
Datatilsynet.
\newblock Offentlige myndigheders brug af kunstig intelligens.
\newblock 2023.
\newblock URL \url{https://www.datatilsynet.dk/Media/638321084132236143/Offentlige%20myndigheders%20brug%20af%20kunstig%20intelligens%20-%20Inden%20I%20går%20i%20gang.pdf}.

\bibitem[de~Vries(2023)]{de2023growing}
de~Vries, A.
\newblock The growing energy footprint of artificial intelligence.
\newblock \emph{Joule}, 7\penalty0 (10):\penalty0 2191--2194, 2023.

\bibitem[Duan et~al.(2024)Duan, Suri, Mireshghallah, Min, Shi, Zettlemoyer, Tsvetkov, Choi, Evans, and Hajishirzi]{duan2024membership}
Duan, M., Suri, A., Mireshghallah, N., Min, S., Shi, W., Zettlemoyer, L., Tsvetkov, Y., Choi, Y., Evans, D., and Hajishirzi, H.
\newblock Do membership inference attacks work on large language models?
\newblock \emph{arXiv preprint arXiv:2402.07841}, 2024.

\bibitem[Dwork(2006)]{dwork2006differential}
Dwork, C.
\newblock Differential privacy.
\newblock In \emph{International colloquium on automata, languages, and programming}, pp.\  1--12. Springer, 2006.

\bibitem[EDPB(2024)]{edpb2024opinion}
EDPB.
\newblock Opinion 28/2024 on certain data protection aspects related to the processing of personal data in the context of ai models.
\newblock 2024.

\bibitem[Elazar et~al.(2021)Elazar, Kassner, Ravfogel, Ravichander, Hovy, Sch{\"u}tze, and Goldberg]{elazar2021measuring}
Elazar, Y., Kassner, N., Ravfogel, S., Ravichander, A., Hovy, E., Sch{\"u}tze, H., and Goldberg, Y.
\newblock Measuring and improving consistency in pretrained language models.
\newblock \emph{Transactions of the Association for Computational Linguistics}, 9:\penalty0 1012--1031, 2021.

\bibitem[{European Data Protection Supervisor}(2019)]{EDPRS}
{European Data Protection Supervisor}.
\newblock Data protection impact assessment list.
\newblock 2019.
\newblock URL \url{https://www.edps.europa.eu/data-protection/our-work/publications/guidelines/data-protection-impact-assessment-list_en}.

\bibitem[Fabbrini \& Celeste(2020)Fabbrini and Celeste]{fabbrini2020right}
Fabbrini, F. and Celeste, E.
\newblock The right to be forgotten in the digital age: The challenges of data protection beyond borders.
\newblock \emph{German law journal}, 21\penalty0 (S1):\penalty0 55--65, 2020.

\bibitem[Feiler et~al.(2018)Feiler, Forg{\'o}, and Weigl]{feiler2018eu}
Feiler, L., Forg{\'o}, N., and Weigl, M.
\newblock The eu general data protection regulation (gdpr): a commentary.
\newblock \emph{(No Title)}, 2018.

\bibitem[Feldman \& Zhang(2020)Feldman and Zhang]{feldman2020neural}
Feldman, V. and Zhang, C.
\newblock What neural networks memorize and why: Discovering the long tail via influence estimation.
\newblock \emph{Advances in Neural Information Processing Systems}, 33:\penalty0 2881--2891, 2020.

\bibitem[Finck \& Pallas(2020)Finck and Pallas]{finck2020they}
Finck, M. and Pallas, F.
\newblock They who must not be identified—distinguishing personal from non-personal data under the gdpr.
\newblock \emph{International Data Privacy Law}, 10\penalty0 (1):\penalty0 11--36, 2020.

\bibitem[Gao et~al.(2020)Gao, Biderman, Black, Golding, Hoppe, Foster, Phang, He, Thite, Nabeshima, et~al.]{gao2020pile}
Gao, L., Biderman, S., Black, S., Golding, L., Hoppe, T., Foster, C., Phang, J., He, H., Thite, A., Nabeshima, N., et~al.
\newblock The pile: An 800gb dataset of diverse text for language modeling.
\newblock \emph{arXiv preprint arXiv:2101.00027}, 2020.

\bibitem[{GDPR ET}(2024)]{AllFine}
{GDPR ET}, G., 2024.
\newblock URL \url{https://www.enforcementtracker.com/}.

\bibitem[Hacker(2021)]{hacker2021legal}
Hacker, P.
\newblock A legal framework for ai training data—from first principles to the artificial intelligence act.
\newblock \emph{Law, innovation and technology}, 13\penalty0 (2):\penalty0 257--301, 2021.

\bibitem[Hacker et~al.(2023)Hacker, Engel, and Mauer]{hacker2023regulating}
Hacker, P., Engel, A., and Mauer, M.
\newblock Regulating chatgpt and other large generative ai models.
\newblock In \emph{Proceedings of the 2023 ACM Conference on Fairness, Accountability, and Transparency}, pp.\  1112--1123, 2023.

\bibitem[Hamburg(2024)]{Hamburg2024LLM}
Hamburg, L.
\newblock Diskussionspapier: Large language models und personenbezogene daten.
\newblock 2024.
\newblock URL \url{https://datenschutz-hamburg.de/fileadmin/user_upload/HmbBfDI/Datenschutz/Informationen/240715_Diskussionspapier_HmbBfDI_KI_Modelle.pdf}.

\bibitem[Hartmann et~al.(2023)Hartmann, Suri, Bindschaedler, Evans, Tople, and West]{hartmann2023sok}
Hartmann, V., Suri, A., Bindschaedler, V., Evans, D., Tople, S., and West, R.
\newblock Sok: Memorization in general-purpose large language models.
\newblock \emph{arXiv preprint arXiv:2310.18362}, 2023.

\bibitem[Hintze(2018)]{hintze2018data}
Hintze, M.
\newblock Data controllers, data processors, and the growing use of connected products in the enterprise: Managing risks, understanding benefits, and complying with the gdpr.
\newblock \emph{Journal of Internet Law (Wolters Kluwer), August}, 2018.

\bibitem[Holzel(2019)]{holzel2019differential}
Holzel, J.
\newblock Differential privacy and the gdpr.
\newblock \emph{Eur. Data Prot. L. Rev.}, 5:\penalty0 184, 2019.

\bibitem[Hu et~al.(2022)Hu, Salcic, Sun, Dobbie, Yu, and Zhang]{hu2022membership}
Hu, H., Salcic, Z., Sun, L., Dobbie, G., Yu, P.~S., and Zhang, X.
\newblock Membership inference attacks on machine learning: A survey.
\newblock \emph{ACM Computing Surveys (CSUR)}, 54\penalty0 (11s):\penalty0 1--37, 2022.

\bibitem[Hu et~al.(2023)Hu, Chen, Li, Guo, Wen, Yu, and Guo]{hu2023large}
Hu, X., Chen, J., Li, X., Guo, Y., Wen, L., Yu, P.~S., and Guo, Z.
\newblock Do large language models know about facts?
\newblock \emph{arXiv preprint arXiv:2310.05177}, 2023.

\bibitem[Huang et~al.(2022)Huang, Shao, and Chang]{huang2022large}
Huang, J., Shao, H., and Chang, K. C.-C.
\newblock Are large pre-trained language models leaking your personal information?
\newblock \emph{arXiv preprint arXiv:2205.12628}, 2022.

\bibitem[Huang et~al.(2024)Huang, Yang, and Potts]{huang2024demystifying}
Huang, J., Yang, D., and Potts, C.
\newblock Demystifying verbatim memorization in large language models.
\newblock \emph{arXiv preprint arXiv:2407.17817}, 2024.

\bibitem[Ippolito et~al.(2022)Ippolito, Tram{\`e}r, Nasr, Zhang, Jagielski, Lee, Choquette-Choo, and Carlini]{ippolito2022preventing}
Ippolito, D., Tram{\`e}r, F., Nasr, M., Zhang, C., Jagielski, M., Lee, K., Choquette-Choo, C.~A., and Carlini, N.
\newblock Preventing verbatim memorization in language models gives a false sense of privacy.
\newblock \emph{arXiv preprint arXiv:2210.17546}, 2022.

\bibitem[Jones \& Kaminski(2020)Jones and Kaminski]{jones2020american}
Jones, M.~L. and Kaminski, M.~E.
\newblock An american's guide to the gdpr.
\newblock \emph{Denv. L. Rev.}, 98:\penalty0 93, 2020.

\bibitem[Juliussen et~al.(2023)Juliussen, Rui, and Johansen]{juliussen2023algorithms}
Juliussen, B.~A., Rui, J.~P., and Johansen, D.
\newblock Algorithms that forget: Machine unlearning and the right to erasure.
\newblock \emph{Computer Law \& Security Review}, 51:\penalty0 105885, 2023.

\bibitem[Kaplan et~al.(2020)Kaplan, McCandlish, Henighan, Brown, Chess, Child, Gray, Radford, Wu, and Amodei]{kaplan2020scaling}
Kaplan, J., McCandlish, S., Henighan, T., Brown, T.~B., Chess, B., Child, R., Gray, S., Radford, A., Wu, J., and Amodei, D.
\newblock Scaling laws for neural language models.
\newblock \emph{arXiv preprint arXiv:2001.08361}, 2020.

\bibitem[Kim et~al.(2024)Kim, Yun, Lee, Gubri, Yoon, and Oh]{kim2024propile}
Kim, S., Yun, S., Lee, H., Gubri, M., Yoon, S., and Oh, S.~J.
\newblock Propile: Probing privacy leakage in large language models.
\newblock \emph{Advances in Neural Information Processing Systems}, 36, 2024.

\bibitem[Kindt et~al.(2021)Kindt, Fontanillo~L{\'o}pez, Bergholm, Czarnocki, Kanevskaia, and Herveg]{kindt2021study}
Kindt, E., Fontanillo~L{\'o}pez, C.~A., Bergholm, J., Czarnocki, J., Kanevskaia, O., and Herveg, J.
\newblock Study on the appropriate safeguards under article 89 (1) gdpr for the processing of personal data for scientific research.
\newblock 2021.

\bibitem[Kloza et~al.(2019)Kloza, Van~Dijk, Casiraghi, Vazquez~Maymir, Roda, Tanas, and Konstantinou]{kloza2019towards}
Kloza, D., Van~Dijk, N., Casiraghi, S., Vazquez~Maymir, S., Roda, S., Tanas, A., and Konstantinou, I.
\newblock Towards a method for data protection impact assessment: Making sense of gdpr requirements.
\newblock \emph{Policy Brief D. Pia. Lab}, 1:\penalty0 1--8, 2019.

\bibitem[Lee et~al.(2022)Lee, Ippolito, Nystrom, Zhang, Eck, Callison-Burch, and Carlini]{lee-etal-2022-deduplicating}
Lee, K., Ippolito, D., Nystrom, A., Zhang, C., Eck, D., Callison-Burch, C., and Carlini, N.
\newblock Deduplicating training data makes language models better.
\newblock In Muresan, S., Nakov, P., and Villavicencio, A. (eds.), \emph{Proceedings of the 60th Annual Meeting of the Association for Computational Linguistics (Volume 1: Long Papers)}, pp.\  8424--8445, Dublin, Ireland, May 2022. Association for Computational Linguistics.
\newblock \doi{10.18653/v1/2022.acl-long.577}.
\newblock URL \url{https://aclanthology.org/2022.acl-long.577/}.

\bibitem[Leiser \& Dechesne(2020)Leiser and Dechesne]{leiser2020governing}
Leiser, M. and Dechesne, F.
\newblock Governing machine-learning models: challenging the personal data presumption.
\newblock \emph{International data privacy law}, 10\penalty0 (3):\penalty0 187--200, 2020.

\bibitem[Li et~al.(2023)Li, Chen, Luo, Wang, Peng, Kang, Zhang, Hu, Chan, Xu, et~al.]{li2023privacy}
Li, H., Chen, Y., Luo, J., Wang, J., Peng, H., Kang, Y., Zhang, X., Hu, Q., Chan, C., Xu, Z., et~al.
\newblock Privacy in large language models: Attacks, defenses and future directions.
\newblock \emph{arXiv preprint arXiv:2310.10383}, 2023.

\bibitem[Lu et~al.(2024)Lu, Li, Cheng, Ding, Huang, and Qiu]{lu2024scaling}
Lu, X., Li, X., Cheng, Q., Ding, K., Huang, X., and Qiu, X.
\newblock Scaling laws for fact memorization of large language models.
\newblock \emph{arXiv preprint arXiv:2406.15720}, 2024.

\bibitem[Meng et~al.(2022)Meng, Sharma, Andonian, Belinkov, and Bau]{meng2022mass}
Meng, K., Sharma, A.~S., Andonian, A., Belinkov, Y., and Bau, D.
\newblock Mass-editing memory in a transformer.
\newblock \emph{arXiv preprint arXiv:2210.07229}, 2022.

\bibitem[Mitrou(2018)]{mitrou2018data}
Mitrou, L.
\newblock Data protection, artificial intelligence and cognitive services: is the general data protection regulation (gdpr)‘artificial intelligence-proof’?
\newblock \emph{Artificial Intelligence and Cognitive Services: Is the General Data Protection Regulation (GDPR)‘Artificial Intelligence-Proof}, 2018.

\bibitem[Nasr et~al.(2023)Nasr, Carlini, Hayase, Jagielski, Cooper, Ippolito, Choquette-Choo, Wallace, Tram{\`e}r, and Lee]{nasr2023scalable}
Nasr, M., Carlini, N., Hayase, J., Jagielski, M., Cooper, A.~F., Ippolito, D., Choquette-Choo, C.~A., Wallace, E., Tram{\`e}r, F., and Lee, K.
\newblock Scalable extraction of training data from (production) language models.
\newblock \emph{arXiv preprint arXiv:2311.17035}, 2023.

\bibitem[Nguyen et~al.(2022)Nguyen, Huynh, Ren, Nguyen, Liew, Yin, and Nguyen]{nguyen2022survey}
Nguyen, T.~T., Huynh, T.~T., Ren, Z., Nguyen, P.~L., Liew, A. W.-C., Yin, H., and Nguyen, Q. V.~H.
\newblock A survey of machine unlearning.
\newblock \emph{arXiv preprint arXiv:2209.02299}, 2022.

\bibitem[NOBY(4)]{Noby2024LLM}
NOBY.
\newblock Chatgpt provides false information about people, and openai can’t correct it.
\newblock 4.
\newblock URL \url{https://noyb.eu/en/chatgpt-provides-false-/information-about-people-and-openai-cant-correct-it}.

\bibitem[Ouyang et~al.(2023)Ouyang, Zhang, Harman, and Wang]{ouyang2023llm}
Ouyang, S., Zhang, J.~M., Harman, M., and Wang, M.
\newblock Llm is like a box of chocolates: the non-determinism of chatgpt in code generation.
\newblock \emph{arXiv preprint arXiv:2308.02828}, 2023.

\bibitem[Peng et~al.(2023)Peng, Galley, He, Cheng, Xie, Hu, Huang, Liden, Yu, Chen, et~al.]{peng2023check}
Peng, B., Galley, M., He, P., Cheng, H., Xie, Y., Hu, Y., Huang, Q., Liden, L., Yu, Z., Chen, W., et~al.
\newblock Check your facts and try again: Improving large language models with external knowledge and automated feedback.
\newblock \emph{arXiv preprint arXiv:2302.12813}, 2023.

\bibitem[Perez et~al.(2022)Perez, Huang, Song, Cai, Ring, Aslanides, Glaese, McAleese, and Irving]{perez2022red}
Perez, E., Huang, S., Song, F., Cai, T., Ring, R., Aslanides, J., Glaese, A., McAleese, N., and Irving, G.
\newblock Red teaming language models with language models.
\newblock \emph{arXiv preprint arXiv:2202.03286}, 2022.

\bibitem[Petroni et~al.(2019)Petroni, Rockt{\"a}schel, Lewis, Bakhtin, Wu, Miller, and Riedel]{petroni2019language}
Petroni, F., Rockt{\"a}schel, T., Lewis, P., Bakhtin, A., Wu, Y., Miller, A.~H., and Riedel, S.
\newblock Language models as knowledge bases?
\newblock \emph{arXiv preprint arXiv:1909.01066}, 2019.

\bibitem[Pope(2024)]{HLR2024LLM}
Pope, A.
\newblock Nyt v. openai: The times’s about-face.
\newblock \emph{Harvard Law Review}, 2024.
\newblock URL \url{https://harvardlawreview.org/blog/2024/04/nyt-v-openai-the-timess-about-face/}.

\bibitem[Power et~al.(2022)Power, Burda, Edwards, Babuschkin, and Misra]{power2022grokking}
Power, A., Burda, Y., Edwards, H., Babuschkin, I., and Misra, V.
\newblock Grokking: Generalization beyond overfitting on small algorithmic datasets.
\newblock \emph{arXiv preprint arXiv:2201.02177}, 2022.

\bibitem[Purtova(2018)]{purtova2018law}
Purtova, N.
\newblock The law of everything. broad concept of personal data and future of eu data protection law.
\newblock \emph{Law, Innovation and Technology}, 10\penalty0 (1):\penalty0 40--81, 2018.

\bibitem[Research \& Innovation(2024)Research and Innovation]{UKRI}
Research, U. and Innovation.
\newblock Gdpr and research – an overview for researchers, 2024.
\newblock URL \url{https://www.ukri.org/wp-content/uploads/2020/10/UKRI-020920-GDPR-FAQs.pdf}.

\bibitem[Roberts et~al.(2020)Roberts, Raffel, and Shazeer]{roberts2020much}
Roberts, A., Raffel, C., and Shazeer, N.
\newblock How much knowledge can you pack into the parameters of a language model?
\newblock \emph{arXiv preprint arXiv:2002.08910}, 2020.

\bibitem[Ruohonen \& Hjerppe(2022)Ruohonen and Hjerppe]{ruohonen2022gdpr}
Ruohonen, J. and Hjerppe, K.
\newblock The gdpr enforcement fines at glance.
\newblock \emph{Information Systems}, 106:\penalty0 101876, 2022.

\bibitem[Sag(2023)]{sag2023fairness}
Sag, M.
\newblock Fairness and fair use in generative ai.
\newblock \emph{Fordham Law Review, Forthcoming}, 2023.

\bibitem[Sartor et~al.(2020)Sartor, Lagioia, et~al.]{sartor2020impact}
Sartor, G., Lagioia, F., et~al.
\newblock The impact of the general data protection regulation (gdpr) on artificial intelligence.
\newblock 2020.

\bibitem[Shi et~al.(2023)Shi, Ajith, Xia, Huang, Liu, Blevins, Chen, and Zettlemoyer]{shi2023detecting}
Shi, W., Ajith, A., Xia, M., Huang, Y., Liu, D., Blevins, T., Chen, D., and Zettlemoyer, L.
\newblock Detecting pretraining data from large language models.
\newblock \emph{arXiv preprint arXiv:2310.16789}, 2023.

\bibitem[Solove(2024)]{solove2024artificial}
Solove, D.~J.
\newblock Artificial intelligence and privacy.
\newblock \emph{Available at SSRN}, 2024.

\bibitem[Somepalli et~al.(2023)Somepalli, Singla, Goldblum, Geiping, and Goldstein]{somepalli2023understanding}
Somepalli, G., Singla, V., Goldblum, M., Geiping, J., and Goldstein, T.
\newblock Understanding and mitigating copying in diffusion models.
\newblock \emph{Advances in Neural Information Processing Systems}, 36:\penalty0 47783--47803, 2023.

\bibitem[Tirumala et~al.(2022)Tirumala, Markosyan, Zettlemoyer, and Aghajanyan]{tirumala2022memorization}
Tirumala, K., Markosyan, A., Zettlemoyer, L., and Aghajanyan, A.
\newblock Memorization without overfitting: Analyzing the training dynamics of large language models.
\newblock \emph{Advances in Neural Information Processing Systems}, 35:\penalty0 38274--38290, 2022.

\bibitem[van~den Burg \& Williams(2021)van~den Burg and Williams]{van2021memorization}
van~den Burg, G. and Williams, C.
\newblock On memorization in probabilistic deep generative models.
\newblock \emph{Advances in Neural Information Processing Systems}, 34:\penalty0 27916--27928, 2021.

\bibitem[Veale et~al.(2018)Veale, Binns, and Edwards]{veale2018algorithms}
Veale, M., Binns, R., and Edwards, L.
\newblock Algorithms that remember: model inversion attacks and data protection law.
\newblock \emph{Philosophical Transactions of the Royal Society A: Mathematical, Physical and Engineering Sciences}, 376\penalty0 (2133):\penalty0 20180083, 2018.

\bibitem[Verma et~al.(2024)Verma, Rassin, Das, Bhatt, Seshadri, Shah, Bilmes, Hajishirzi, and Elazar]{verma2024many}
Verma, S., Rassin, R., Das, A., Bhatt, G., Seshadri, P., Shah, C., Bilmes, J., Hajishirzi, H., and Elazar, Y.
\newblock How many van goghs does it take to van gogh? finding the imitation threshold.
\newblock \emph{arXiv preprint arXiv:2410.15002}, 2024.

\bibitem[Villalobos et~al.(2024)Villalobos, Ho, Sevilla, Besiroglu, Heim, and Hobbhahn]{villalobos2024will}
Villalobos, P., Ho, A., Sevilla, J., Besiroglu, T., Heim, L., and Hobbhahn, M.
\newblock Will we run out of data? limits of llm scaling based on human-generated data.
\newblock \emph{arXiv preprint arXiv:2211.04325}, pp.\  13--29, 2024.

\bibitem[Villaronga et~al.(2018)Villaronga, Kieseberg, and Li]{villaronga2018humans}
Villaronga, E.~F., Kieseberg, P., and Li, T.
\newblock Humans forget, machines remember: Artificial intelligence and the right to be forgotten.
\newblock \emph{Computer Law \& Security Review}, 34\penalty0 (2):\penalty0 304--313, 2018.

\bibitem[Vrabec(2021)]{vrabec2021data}
Vrabec, H.~U.
\newblock \emph{Data Subject Rights under the GDPR}.
\newblock Oxford University Press, 2021.

\bibitem[Wachter \& Mittelstadt(2019)Wachter and Mittelstadt]{wachter2019right}
Wachter, S. and Mittelstadt, B.
\newblock A right to reasonable inferences: re-thinking data protection law in the age of big data and ai.
\newblock \emph{Colum. Bus. L. Rev.}, pp.\  494, 2019.

\bibitem[Wachter et~al.(2017)Wachter, Mittelstadt, and Russell]{wachter2017counterfactual}
Wachter, S., Mittelstadt, B., and Russell, C.
\newblock Counterfactual explanations without opening the black box: Automated decisions and the gdpr.
\newblock \emph{Harv. JL \& Tech.}, 31:\penalty0 841, 2017.

\bibitem[Wang et~al.(2023)Wang, Liu, Yue, Tang, Zhang, Jiayang, Yao, Gao, Hu, Qi, et~al.]{wang2023survey}
Wang, C., Liu, X., Yue, Y., Tang, X., Zhang, T., Jiayang, C., Yao, Y., Gao, W., Hu, X., Qi, Z., et~al.
\newblock Survey on factuality in large language models: Knowledge, retrieval and domain-specificity.
\newblock \emph{arXiv preprint arXiv:2310.07521}, 2023.

\bibitem[Wolters(2018)]{wolters2018control}
Wolters, P.
\newblock The control by and rights of the data subject under the gdpr.
\newblock 2018.

\bibitem[Xu et~al.(2024)Xu, Wu, Wang, and Jia]{xu2024machine}
Xu, J., Wu, Z., Wang, C., and Jia, X.
\newblock Machine unlearning: Solutions and challenges.
\newblock \emph{IEEE Transactions on Emerging Topics in Computational Intelligence}, 2024.

\bibitem[Yaish(2019)]{yaish2019forget}
Yaish, H.
\newblock Forget me, forget me not: Elements of erasure to determine the sufficiency of a gdpr article 17 request.
\newblock \emph{Case W. Res. JL Tech. \& Internet}, 10:\penalty0 iv, 2019.

\bibitem[Yao et~al.(2024)Yao, Duan, Xu, Cai, Sun, and Zhang]{yao2024survey}
Yao, Y., Duan, J., Xu, K., Cai, Y., Sun, Z., and Zhang, Y.
\newblock A survey on large language model (llm) security and privacy: The good, the bad, and the ugly.
\newblock \emph{High-Confidence Computing}, pp.\  100211, 2024.

\bibitem[Yu et~al.(2022)Yu, Iter, Wang, Xu, Ju, Sanyal, Zhu, Zeng, and Jiang]{yu2022generate}
Yu, W., Iter, D., Wang, S., Xu, Y., Ju, M., Sanyal, S., Zhu, C., Zeng, M., and Jiang, M.
\newblock Generate rather than retrieve: Large language models are strong context generators.
\newblock \emph{arXiv preprint arXiv:2209.10063}, 2022.

\bibitem[Zhang et~al.(2023{\natexlab{a}})Zhang, Ippolito, Lee, Jagielski, Tram{\`e}r, and Carlini]{zhang2023counterfactual}
Zhang, C., Ippolito, D., Lee, K., Jagielski, M., Tram{\`e}r, F., and Carlini, N.
\newblock Counterfactual memorization in neural language models.
\newblock \emph{Advances in Neural Information Processing Systems}, 36:\penalty0 39321--39362, 2023{\natexlab{a}}.

\bibitem[Zhang et~al.(2023{\natexlab{b}})Zhang, Nakamura, Isohara, and Sakurai]{zhang2023review}
Zhang, H., Nakamura, T., Isohara, T., and Sakurai, K.
\newblock A review on machine unlearning.
\newblock \emph{SN Computer Science}, 4\penalty0 (4):\penalty0 337, 2023{\natexlab{b}}.

\end{thebibliography}

\bibliographystyle{icml2025}

\appendix
\section{Appendix}

\begin{figure} [h!]
    \centering
    \includegraphics[width=1\linewidth]{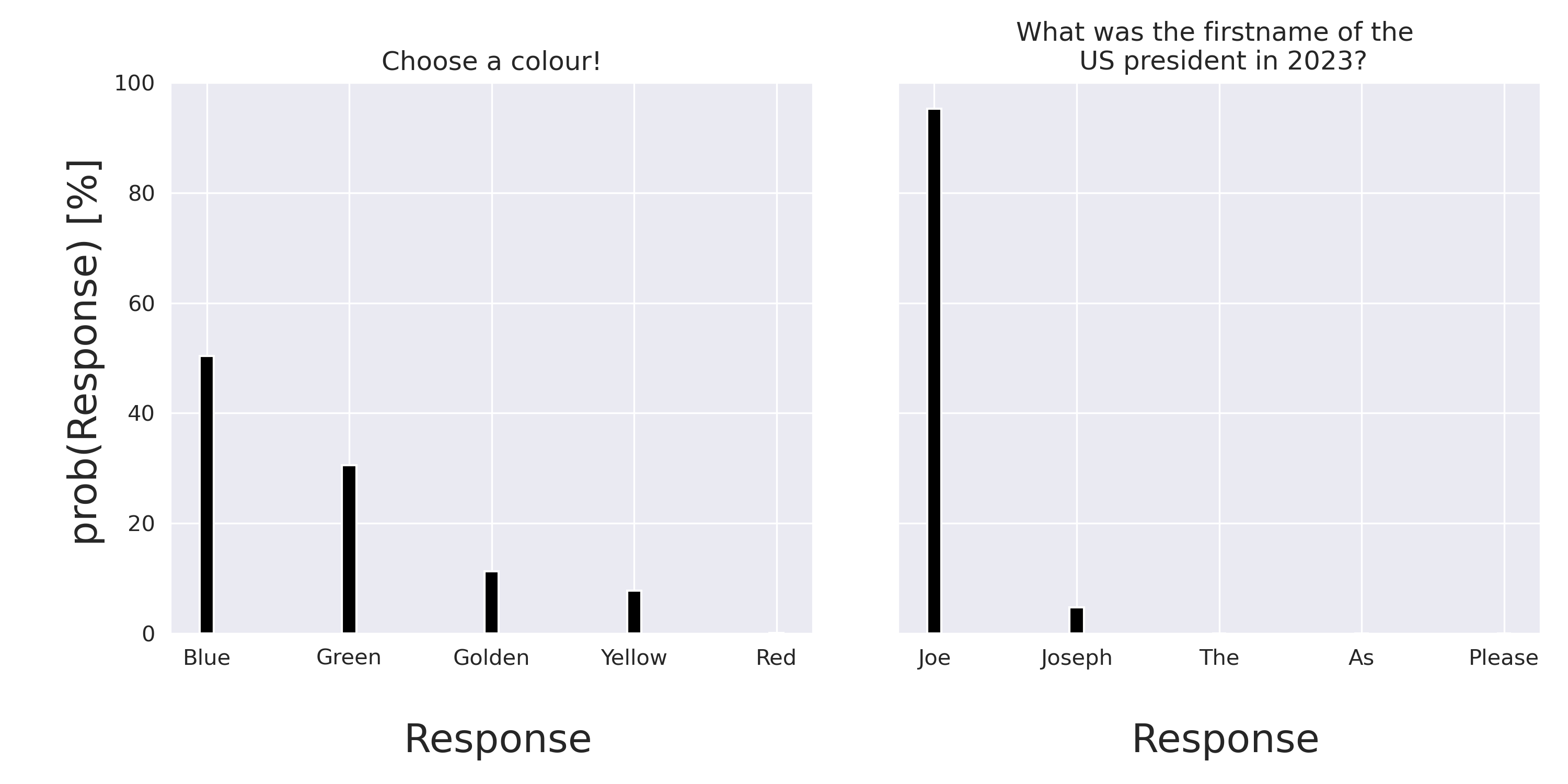}
    \caption{Large Language Models are always probabilistic. (a) When GPT-4o is asked to choose a color, we observe probabilistic behavior. The y-axis shows the probabilities output by the OpenAI API. (b) When asked who was the president of the USA in 2023, the probabilistic behavior is hidden by a high-likelihood single response.}
    \label{fig:Probs}
\end{figure}

\end{document}